%% file: paper.tex
\documentclass[dvipdfmx,10pt]{article}

\usepackage{amsfonts,amsmath,amsthm,amssymb}
\usepackage{url}
\usepackage{bm,color}
\usepackage{graphicx} 
\usepackage[margin=1in]{geometry}
\usepackage{dcolumn,multirow}

\usepackage{mcr,thmE}

\title{An Efficient Post-Selection Inference on \\ High-Order Interaction Models}

\date{June 22, 2015}

\author{
S.~Suzumura\\
Nagoya Institute of Technology\\
\texttt{suzumura.mllab.nit@gmail.com}
\\
\and
K.~Nakagawa\\
Nagoya Institute of Technology\\
\texttt{nakagawa.k.mllab.nit@gmail.com}
\\
\and
K.~Tsuda\\
Tokyo University\\
\texttt{tsuda@k.u-tokyo.ac.jp}
\\
\and
I.~Takeuchi\thanks{Corresponding Author}\\
Nagoya Institute of Technology\\
\texttt{takeuchi.ichiro@nitech.ac.jp}
}

\begin{document}

\maketitle

\begin{abstract}
\input{abst}

\end{abstract}

\input{sec1}
\input{sec2}
\input{sec3}
\input{sec4}

\input{sec5}
\newpage
\bibliographystyle{unsrt}

\appendix

\clearpage
\input{appA}

\end{document}

%% file: abst.tex
Finding statistically significant high-order interaction features in predictive modeling
is important but challenging task.
The difficulty lies in the fact that,
for a recent applications with high-dimensional covariates, 
the number of possible high-order interaction features would be extremely large.
Identifying statistically significant features from such a huge pool of candidates
would be highly challenging both in computational and statistical senses. 
To work with this problem, 
we consider a two stage algorithm 
where
we first select a set of high-order interaction features
by marginal screening,
and then make statistical inferences on 
the regression model fitted only with the selected features. 
Such statistical inferences are called 
\emph{post-selection inference (PSI)},
and 
receiving an increasing attention
in the literature. 
One of the seminal recent advancements
in PSI literature 
is the works by Lee et al. 
\cite{Lee14a,Lee15a},
where
the authors presented an algorithmic framework 
for computing exact sampling distributions in PSI. 
A main challenge when applying their approach to our high-order interaction models 
is to cope with the fact that PSI in general 
depends not only on the selected features but also on the unselected features, 
making it hard to apply to our extremely high-dimensional high-order interaction models. 
The goal of this paper is to overcome this difficulty 
by introducing a novel efficient method for PSI.
Our key idea is to 
exploit the underlying tree structure among 
high-order interaction features,
and to develop a pruning method of the tree 
which enables us to quickly identify a group of unselected features 
that are guaranteed to have no influence on PSI. 
The experimental results indicate that
the proposed method allows us
to reliably identify statistically significant high-order interaction features
with reasonable computational cost.

%% file: sec1.tex
\section{Introduction}
Finding statistically reliable high-order interaction features 
that have  significant effects on the response
is valuable 
in many regression problems. 
For example,
in biomedical studies,
it is well-known that 
each genetic factor
such as a single gene 
does not work independently.
When a regression analysis is used 
in biomedical studies 
for predicting a certain phenotype
such as drug response, 
high-order interactions of multiple genetic factors
might be useful \cite{manolio2006genes, cordell2009detecting}.
%
%
If one has a data set with $d$ original covariates
and
takes into account interaction terms 
up to order $r$, 
the regression model has 
$\scriptstyle D := \sum_{\rho = 1}^r {d \choose \rho}$
features. 
Unless both $d$ and $r$ are fairly small,
the number of features
$D$
would be far greater than the sample size
$n$.
Statistical inferences 
on such an extremely high-dimensional regression model 
is quite challenging. 

A common practical approach to high-dimensional regression problems
is
\emph{two-stage method},
where 
a subset of features is first selected, 
and then a regression model only with the selected features is fitted. 
A statistical issue of such a two-stage method is 
how to incorporate the effect of the feature selection stage
on the statistical inference of the final regression model. 
If the two stages 
are performed with the same data set, 
confidence intervals or $p$-values
on the final regression model 
would be positively biased.
Statistical inferences conditional on \emph{pre}-feature selection 
is often called
\emph{post-selection inference (PSI)}.
Until recently,
PSI has been recognized to be intractable
in most cases 
because 
it seems to be difficult to derive the sampling distribution
that can fully account
for complex feature selection process \cite{leeb2005model,leeb2006can}.
Recently, 
Lee et al. \cite{Lee14a,Lee15a}
introduced
an affirmative solution to PSI
for a wide class of feature selection methods.
Specifically,
they provided a general algorithm 
for computing an exact sampling distribution of the response 
conditional on a feature selection event 
which is represented
by a set of affine constraints
in the response domain.
A notable advantage of their finding is that 
many commonly used feature selection algorithms 
such as 
marginal screening,
orthogonal matching pursuit, 
and 
Lasso 
belong to this class.
%
%
Using the sampling distribution of the response conditional on a feature selection event, 
one can make various statistical inferences on the \emph{post}-regression model 
that
properly incorporate the effect of \emph{pre}-feature selection. 

The goal of this paper is
to develop a method
for finding statistically significant high-order interaction features
by using the idea of Lee et al. \cite{Lee14a,Lee15a}. 
Unfortunately,
their method 
cannot be directly applied to 
our extremely high-dimensional regression model with high-order interaction features. 
The difficulty lies in the simple fact that 
a feature selection event
in general
depends 
not only on the selected features but also on the unselected features. 
It suggests that,
at least 
$\cO(D)$
constraints would be needed for characterizing a feature selection event. 
Since the number of features $D$ is extremely large
in our high-order interaction model,
it would be computationally intractable to work with 
all those constrains. 
In this paper
we mainly study PSI on high-order interaction models
with marginal screening-based pre-feature selection.
In marginal screening, 
we select  top $k$ features
from all the $D$ features 
according to the association of each feature with the response. 
Despite its simplicity, 
marginal screening is one of the most-frequently used feature selection methods,
and
it has been shown to have several desirable statistical properties
under some regularity conditions 
\cite{Fan08a,Fan09a,Fan10a,Genovese12a}. 
%
%
%
As we describe in the next section,
a feature selection event by marginal screening is
characterized by a set of
$2k\bar{k} + k$
affine constraints in the response domain,
where
$\bar{k}: = D - k$
is the number of unselected features. 
It suggests that 
the sampling distribution of the response
conditional on the marginal screening 
depends
in general
on these
$2k\bar{k} + k$
affine constraints.

Our main contribution in this paper is to develop a novel algorithm
that can efficiently find a subset of these
$2k\bar{k} + k$
affine constraints 
which are guaranteed to have no influence on the conditional sampling distribution. 
Our basic idea is to exploit the underlying tree structure
among a set of high-order interaction features
(see \figurename~\ref{fig:tree}).
Specifically,
we derive an efficient pruning condition of the tree
such that,
for any node in the tree,
if a certain condition on the node is satisfied,
then all the features corresponding to its descendant nodes are shown to have no influence
on the conditional sampling distribution.
As demonstrated in the experiment section, 
our algorithm allows us to work with a PSI for a high-order interaction model
e.g., 
with
$d = 5000$
and
$r = 5$ 
where the number of all the high-order interaction features $D$ is greater than $10^{16}$.

%% file: sec2.tex
\section{Preliminaries}
\label{sec:preliminaries}
%

\paragraph{Problem setup}
Consider
modeling a relationship
between a response 
$Y \in \RR$
and
$d$-dimensional covariates
$\bm z = [z_1, \ldots, z_d]^\top$
by the following high-order interaction model up to $r^{\rm th}$ order
\begin{align}
 \label{eq:homdl}
 Y
 =
 \sum_{j_1 \in [d]} \alpha_{j_1} z_{j_1}
 +
 \sum_{\substack{
 (j_1, j_2) \in [d] \times [d] \\
 j_1 \neq j_2
 }
 } \alpha_{j_1, j_2} z_{j_1} z_{j_2}
 +
 \ldots
 +
 \sum_{\substack{(j_1, \ldots, j_r) \in [d]^r \\ j_1 \neq \ldots \neq j_r}}
 \alpha_{j_1, \ldots, j_r} z_{j_1} \cdots z_{j_r}
 +
 \veps, 
\end{align}
where
$\alpha$s
are the coefficients 
and 
$\veps$
is a random noise. 
%
We assume that each original covariate 
$z_j, j \in [d]$
is defined in a domain $[0, 1]$
where
values 1 and 0 respectively indicate the existence and the non-existence of a certain property,
and values between them indicate the ``degree'' of existence. 
High-order interaction features thus represent co-existence of multiple properties.
For example,
if
$z_{j_1}$
represents high body mass index (BMI)
and 
$z_{j_2}$
represents a mutation in a certain gene,
we may code these two covariates as
\begin{align*}
 z_{j_1} := \mycase{
 1 & \text{ if BMI  } > 30, \\
 (\text{BMI} - 15)/(30 - 15) & \text{ if BMI } \in [15, 30], \\
 0 & \text{ if BMI } < 15
 }
 ~
 z_{j_2} := \mycase{
 1 & \text{ if there is a mutation}, \\
 0 & \text{ if there is no mutation}.
 }
\end{align*}
Then,
an interaction term 
$z_{j_1} z_{j_2}$
represents
the co-existence of high BMI and a mutation in the gene. 

The high-order interaction model 
\eq{eq:homdl}
has in total 
$D := \sum_{\rho \in [r]} {d \choose \rho}$
features. 
Let us write the mapping 
from the original covariates 
$\bm z := [z_1, \ldots, z_d]^\top \in \RR^d$
to the high-order interaction features 
$\bm x := [x_1, \ldots, x_D]^\top \in \RR^D$
as
$
\bm \phi: [0, 1]^d \to [0, 1]^D, \bm z \mapsto \bm x,
$
where the latter has defined as 
\begin{align}
\label{eq:expandedX}
 \bm x :=
 \bm \phi(\bm z) = [
z_1, \ldots, z_d,
z_1 z_2, \ldots, z_{d-1} z_d,
\ldots,
z_1 \! \cdot \! \cdot \! \cdot \! z_k,
\ldots,
z_{d-k+1}  \! \cdot \! \cdot \! \cdot \! z_d
]^\top \in \RR^{D}.
\end{align}
%
Since a high-order interaction feature is a product of original covariates defined in $[0, 1]$, 
the range of each feature 
$x_j, j \in [D]$ 
is also $[0, 1]$.

Our goal is to identify statistically significant high-order interaction terms
that have large impacts on the response $Y$ 
by identifying regression coefficients
$\alpha$s 
which are significantly deviated from zero.
However, 
unless both
$d$ and $r$
are fairly small, 
the number of coefficients $\alpha$s 
to be fitted 
would be far greater than the sample size
$n$,
meaning that the unique least-square solution does not exist,
and
traditional least-square estimation theory cannot be used for 
making statistical inferences on the fitted model.
We thus introduce PSI framework
where 
a subset of features is first selected by marginal screening,
and then statistical inferences
on the fitted model only with the selected features are considered. 

\paragraph{Post-selection inference with marginal screening}
In the high-order interaction feature domain
$[0, 1]^D$, 
we consider the same problem setup
as
Lee et al.'s work \cite{Lee14a,Lee15a}. 
We assume that the data is generated from the following process
\begin{align}
 \label{eq:data-generator}
 \bm \cY \sim N(\bm \mu, \bm \Sigma), 
\end{align}
where 
$\bm \cY \in \RR^n$
is a random response vector Normally distributed with the mean vector
$\bm \mu \in \RR^n$
and the variance-covariance matrix 
$\bm \Sigma \in \RR^{n \times n}$.
%
The mean vector
$\bm \mu$
in general depends on the fixed (non-random) design matrix
$\bm X \in [0, 1]^{n \times D}$.
%
The training set is denoted as
$(\bm X, \bm y)$
where
$\bm y \in \RR^n$
is an observed response from the data generating process 
\eq{eq:data-generator}.
The training set is also denoted as
$\{(\bm x^i, y^i)\}_{i \in [n]}$
where
$\bm x^i \in [0, 1]^D$
is the $i^{\rm th}$ row of $\bm X$
and
$y_i \in \RR$
is the $i^{\rm th}$ element of $\bm y$.
Similarly,
the $j^{\rm th}$ column of $\bm X$
is denoted as 
$\bm x_j$
for
$j \in [D]$. 

\paragraph{Marginal screening}
In the first stage,
we select 
top $k$ features that have strong association with the response.
Noting that each feature is defined in
$[0, 1]$
and the value indicates 
(the degree of) the existence of a certain property, 
we consider a score
$\bm x_j^\top \bm y$
for each of the $D$ features, 
and select $k$ top features according to their absolute scores
$\{|\bm x_j^\top \bm y|\}_{j \in [D]}$. 

We denote the index set of the 
selected
$k$
features
by 
$\cS$,
and
that of the unselected
$\bar{k}: =D - k$
features
by 
$\bar{\cS}$. 
As pointed out in \cite{Lee14a},
marginal screening event is characterized by a set of affine constraints.
The fact that
$k$ features in $\cS$ are selected
and
$\bar{k}$ features in $\bar{\cS}$ are not selected
is rephrased by
\begin{align}
 \label{eq:selection-event-1}
 | \bm x_j^\top \bm y |
 \ge
 | \bm x_{\ell}^\top \bm y |
 \text{ for all }
 (j, \ell)
 \in
 \cS \times \bar{\cS}.
\end{align}
Let
$\hat{s}_j := {\rm sign}(\bm x_j^\top \bm y), j \in \cS$. 
Then 
the feature selection event in
\eq{eq:selection-event-1} 
is rewritten
with the sign constraints of the selected features
by the following 
$2 k\bar{k} + k$
constraints
\begin{align}
 \label{eq:selection-event-2}
  (- \hat{s}_j \bm x_j - \bm x_\ell)^\top \bm y \le 0,
  ~
  (- \hat{s}_j \bm x_j + \bm x_\ell)^\top \bm y \le 0,
  ~
 - \hat{s}_j \bm x_j^\top \bm y \le 0
 ~~~
 \forall
  (j, \ell) \in \cS \times \bar{\cS}. 
\end{align}
Since the result of marginal screening depends on the observed response vector
$\bm y$,
we write the feature selection process as a function in the following form
\begin{align*}
 \{\cS, \bar{\cS}, \hat{\bm s}\} = \Omega(\bm y),
\end{align*}
where $\hat{\bm s}$ is denoted as $\hat{s}_j$ for $j \in \cS$.
The set of 
$2 k\bar{k}+k$
constraints 
in \eq{eq:selection-event-2}
is written as
$\bm A \bm y \le \bm b$\footnote{
In the case of marginal screening, 
the vector
$\bm b = \zero$. 
However, 
we keep a vector $\bm b$ here 
for generality: 
if other feature selection method is used
such as Lasso, 
$\bm b \neq \zero$
in general. 
},
for a matrix
$\bm A \in \RR^{(2k\bar{k} + k) \times n}$
and a vector
$\bm b \in \RR^{2k\bar{k} + k}$.

\paragraph{Post-selection inferences}
In the second stage, 
we consider a linear regression model only with the selected features.
Let
$\bm X_{\cS} \in [0, 1]^{n \times k}$ 
be a submatrix of 
$\bm X$
whose columns are indexed by $\cS$.
The best linear unbiased estimator of the regression coefficients 
is the following least-square estimator
\begin{align}
 \label{eq:lin-mdl-selected}
 \hat{\bm \beta}_{\cS}
 :=
 \bm X_{\cS}^{\dagger \top} \bm y,
 \text{ where }
 \bm X_{\cS}^{\dagger} := \bm X_{\cS} (\bm X_{\cS}^\top \bm X_{\cS})^{-1}. 
\end{align}
The population counterpart of
\eq{eq:lin-mdl-selected}
is written as
$
\bm \cB_{\cS} := \bm X_{\cS}^{\dagger \top} \bm \cY.
$
Under the data generating process 
\eq{eq:data-generator}, 
the distribution of 
$\bm \cB_{\cS}$
is written as
\begin{align}
 \label{eq:non-post-sampling-dist}
 \bm \cB_{\cS} \sim N
 (
 \bm X_{\cS}^{\dagger \top}
 \bm \mu,
 ~
 \bm X_{\cS}^{\dagger \top}
 \bm \Sigma
 \bm X_{\cS}^{\dagger}
 )
\end{align}

If the set of features
$\cS$
is fixed a priori,
then we can make statistical inferences on 
$\bm \cB_{\cS}$
by using the sampling distribution 
\eq{eq:non-post-sampling-dist}.
However,
if 
$\cS$
is selected
based on
$\bm y$,
the distribution 
\eq{eq:non-post-sampling-dist}
no longer holds. 
In PSI framework,
statistical inferences should be made based on
the distribution of
$\bm \cB_{\cS}$
conditional on the feature selection event
$\{\cS, \bar{\cS}, \hat{\bm s}\} = \Omega(\bm \cY)$
i.e.,
we need to have distributional result of the conditional random variable 
$
 \bm X_{\cS}^{\dagger} \bm \cY
 ~\big|~
 \{\cS, \bar{\cS}, \hat{\bm s}\} = \Omega(\bm \cY). 
$

The following theorem presented by Lee et al.\cite{Lee14a,Lee15a}
enables us to make post-selection inferences 
as long as the pre-feature selection event is characterized by 
a set of affine constraints $\bm A \bm y \le \bm b$.
\begin{theo}[Lee et al.~\cite{Lee14a,Lee15a}]
 \label{theo:LeeEtAl}
 Consider a stochastic data generating process 
 $\bm \cY \sim N(\bm \mu, \bm \Sigma)$.
 If a feature selection event is characterized by
 $\bm A \bm \cY \le \bm b$
 for an arbitrary matrix 
 $\bm A$
 and a vector 
 $\bm b$
 that do not depend on $\bm \cY$,
 then, 
 for any vector $\bm \eta \in \RR^n$,
 \begin{align*}
  F_{\bm \eta^\top \bm \mu, \bm \eta \bm \Sigma \bm \eta}^{[V^{+}(\bm A, \bm b), V^{-}(\bm A, \bm b)]}
  (\bm \eta^\top \bm \cY)
  ~ | ~
  \bm A \bm \cY \le \bm b
  ~\sim~
  {\rm Unif}(0, 1), 
 \end{align*}
 where
 $F_{t, u}^{[v, w]}(\cdot)$
 is the cumulative distribution function of the univariate truncated Normal distribution
 with the mean $t$, the variance $u$,
 and the lower and the upper truncation points
 $v$ and $w$,
 respectively. 
 Furthermore, 
 using
 $\bm c := \frac{\Sigma \bm \eta}{\bm \eta^\top \Sigma \bm \eta}$, 
 the lower and the upper truncation points are given as
 \begin{align*}
  V^-(\bm A, \bm b)
  :=
  \max_{j : (\bm A \bm c)_j < 0}
  \left\{
  \frac{b_j - (\bm A \bm y)_j}{(\bm A \bm c)_j} 
  \right\}
  + \bm \eta^\top \bm y,
  ~
  V^+(\bm A, \bm b)
  :=
  \min_{j : (\bm A \bm c)_j > 0}
  \left\{
  \frac{b_j - (\bm A \bm y)_j}{(\bm A \bm c)_j} 
  \right\}
  + \bm \eta^\top \bm y.
 \end{align*}
\end{theo}

Theorem~\ref{theo:LeeEtAl}
indicates that, 
if we set
$\bm \eta := \bm X_{\cS}^{\dagger} \bm e_j$,
then
the sampling distribution of 
$\cB_{\cS, j} | \{\cS, \bar{\cS}, \hat{\bm s}\} = \Omega(\bm \cY)$
is a truncated Normal,
where
$\bm e_j$
is a vector of all 0 except 1 in the $j^{\rm th}$ position,
and
$\cB_{\cS, j}$
is the
$j^{\rm th}$
element of
$\bm \cB_{\cS}$. 
If the lower truncation point 
$V^-(\bm A, \bm b)$
and the upper truncation point 
$V^+(\bm A, \bm b)$
can be computed,
we can make post-selection inferences on 
each coefficient 
of the final regression model in the second stage.

However, 
we cannot handle all the 
$2k\bar{k} + k$
constraints
in
$\bm A \bm y \le \bm b$
because
$D$ is exponentially large 
in our high-order interaction models. 
In \S\ref{sec:main},
we develop an efficient algorithm
by exploiting the underlying tree structure among a set of high-order interaction features
that enables us to compute the sampling distribution of
$\bm \eta^\top \bm \cY ~|~ \bm A \bm \cY \le \bm b$
even when
$\bm A$
has exponentially large number of rows. 

\paragraph{Related works}
Before presenting our main contribution,
we briefly review related works in the literature.
Methods for efficiently finding high-order interaction features
and
properly evaluating their statistical significances
have long been desired in many practical application domains.
In the past decade, 
several authors studied this topic in the context of sparse learning
\cite{Choi10a, hao2014interaction, Bien13a}. 
These methods cannot be used for statistical inferences on the selected features 
because 
their main focus is on asymptotic feature selection consistency.
In addition, 
none of these works have special computational trick  
for handling exponentially large number of interaction features,
which makes their empirical evaluations restricted to be only up-to second order interactions. 
One commonly used heuristic in the context of interaction modeling is
to introduce a prior knowledge such as 
\emph{strong heredity assumption} \cite{Choi10a, hao2014interaction, Bien13a},
where, e.g.,
an interaction term
$z_1 z_2$
would be selected
only when both of
$z_1$
and
$z_2$
are selected. 
Such a heuristic restriction is helpful for reducing 
the number of interaction terms to be considered.
However,
in many applications, 
scientists are primarily interested in
finding strong interaction features
even when their main effects alone do not have any association with the response. 
The idea of
considering a structure among the features 
and
utilizing some pruning rules 
is common technique in data mining literature \cite{hamalainen2014statistically,Saigo06a,Kudo05a,Morishita02a}.
%
%
Unfortunately,
it is difficult to properly assess the statistical significances of the selected features by these mining techniques.
%

One traditional approach to assessing the statistical properties on pre-selected features 
is \emph{multiple testing correction (MTC)}.
In the context of DNA microarray studies,
many MTC procedures
for high-dimensional data have been proposed \cite{tusher2001significance,dudoit2003multiple}. 
An MTC approach
for statistical evaluation of high-order interaction features
was recently studied in \cite{terada2013statistical,lopez2015fast}. 
A main drawback of MTC 
is that they are highly conservative 
when the number of candidate features increases.
Another common approach
is \emph{data-splitting (DS)}.
%
In DS approach, 
we split the data into two subsets, 
and use one for feature selection and another for model assessment,
which enables us to 
remove
the PSI bias.
However,
the power of DS approach
is
clearly weaker
than 
the PSI framework
by Lee et al. 
because 
only a part of the available sample is used for
statistical model assessment. 
In addition,
it is quite annoying that 
different set of features
could be selected
if data is splitted differently. 
Despite two-stage method is frequently used in practical high-dimensional data analysis, 
proper PSI methods have not been available until recently. 
Besides the approach by Lee et al.\cite{Lee14a,Lee15a},
several new directions to PSI have been studied lately
\cite{berk2013valid,lockhart2014significance}.
The main contribution of this paper is to develop a practical algorithm 
for proper statistical assessment of high-order interaction features 
based on these recent progress on PSI literature. 
%

%% file: sec3.tex
\section{Efficient post-selection inferences for high-order interaction models}
\label{sec:main}
\begin{figure}[t]
 \centering
 \includegraphics[width=0.7\textwidth]{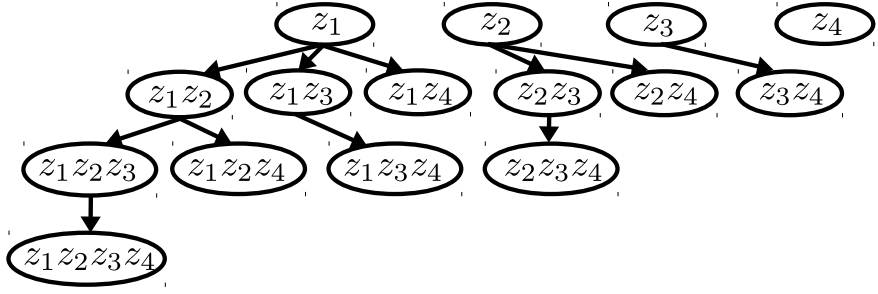} 
 \caption{An underlying tree structure among high-order interaction features ($d = 4, r = 3$).}
 \label{fig:tree}
\end{figure}

In this section we present 
an efficient algorithm
for statistical inferences on high-order interaction model
based on post-selection inference framework.
The basic idea is to exploit the underlying tree structure
among a set of high-order interaction terms
as depicted in
\figurename~\ref{fig:tree}.
Using the tree structure 
we derive a set of pruning conditions of the tree
that allows us to efficiently compute
the sampling distribution 
conditional on the marginal screening.
even when
it is characterized by exponentially large number of affine constraints. 
In what follows, 
for any node $j$ in the tree,
we let 
$De(j)$
be the set of all its descendant nodes.
In
\S~\ref{subsec:efficient-ms},
we describe a simple computational trick 
for marginal screening
when there are exponentially large number of high-order interaction features.
Then,
in
\S~\ref{subsec:efficient-ps},
we present our main results on efficient post-selection inference
for high-order interaction models. 

\subsection{Efficient marginal screening for high-order interaction models}
\label{subsec:efficient-ms}
In the first marginal screening stage,
we select the top 
$k$
features
according to the absolute scores 
$|\bm x_j^\top \bm y|, j \in [D]$. 
In naive implementation,
the absolute scores for all the $D$ features
are first computed,
and then top $k$ of them are selected. 
The computational cost of such a naive implementation is 
$\cO(nD)$,
which is computationally intractable
for our high-order interaction models.
To circumvent the computational cost,
we use the following Lemma. 
\begin{lemm}
 \label{lemm:ms}
 Consider high-order interaction feature vectors 
 $\bm x_j \in [0, 1]^n, j \in [D]$, 
 whose indices are represented in the tree structure
 depicted in 
 \figurename~\ref{fig:tree}.
 Then,
 for any node $j \in [D]$ in the tree, 
 \begin{align}
  \label{eq:lemma:ms}
  | \bm x_{\ell}^\top \bm y |
  \le
  \max\{
  \sum_{i : y^i > 0} x^i_j y^i, 
    - \sum_{i : y^i < 0} x^i_j y^i
  \}
  ~~~\text{ for all }~~~
  \ell \in De(j),
 \end{align}
 where
 $x^i_j$
 is the $(i, j)^{\rm th}$ element of the design matrix
 $\bm X$,
 i.e.,
 the $i^{\rm th}$ element of the vector $\bm x_j$. 
\end{lemm}
This simple lemma can be easily proved by noting that, 
for any
$(i, j) \in [n] \times [D]$, 
$x^i_{\ell} \le x^i_j \in [0, 1]$
for all
$\ell \in De(j)$. 
The lemma has been also used in the context of itemset mining 
\cite{Saigo06a,Kudo05a,Morishita02a}.

Lemma~\ref{lemm:ms}
suggests that 
we can exploit the tree structure 
for efficiently selecting the top $k$ features. 
In depth first search, 
if the left-hand-side of
\eq{eq:lemma:ms}
in a certain node
$j$ 
is smaller than the
$k^{\rm th}$
largest absolute score
obtained so far,
we can quit searching over its descendant nodes
because the lemma indicates that there are no features 
whose absolute score is greater than
the current $k^{\rm th}$ largest one 
in the subtree.
%

\subsection{Efficient post-selection inference for high-order interaction models}
\label{subsec:efficient-ps}
In this section we present our main contribution.
As we saw in \S~\ref{sec:preliminaries},
a marginal-screening event is represented by
$2k\bar{k} + k$
affine constraints.
Theorem~\ref{theo:LeeEtAl}
indicates that
a feature selection event
characterized by such a set of affine constraints
$\bm A \bm y \le \bm b$ 
changes the sampling distribution of the post-regression model 
through
the dependencies of the lower and the upper truncation points 
$V^-(\bm A, \bm b)$
and
$V^+(\bm A, \bm b)$
on the matrix $\bm A$ and the vector $\bm b$.
Our basic idea is to efficiently identify a subset of affine constraints
(a subset of the rows in $\bm A$ and the elements in $\bm b$)
that have no influences on the lower and the upper truncation points
by using a set of pruning conditions
in the tree structure. 


\begin{theo}
 \label{theo:main}
 Let
 $\theta \in [2k\bar{k}]$
 be the index of the first
$2 k \bar{k}$
 affine constraints in
 \eq{eq:selection-event-2},
 and let
 $\cC_1 := \{1, \ldots, k\bar{k} \}$,
and
 $\cC_2 := \{k\bar{k} + 1, \ldots, 2k\bar{k}\}$.
 Furthermore,
 for notational simplicity, 
 assume that first
 $k$
 features are selected and remaining
 $\bar{k} = D - k$
 features are unselected. 
 Then, 
 aside from the sign constraints
 $\hat{s}_j \bm x_j^\top \bm y \ge 0, j \in \cS$,
 a marginal screening event
 $\bm A \bm y \le \bm b$
 in \eq{eq:selection-event-2}
 is written
 as 
 \begin{align*}
  &
  (\phantom{-} \bm x_{\ell(\theta)} - \hat{s}_{j(\theta)} \bm x_{j(\theta)} )^\top \bm y \le 0
  ~\text{ with }~
  j(\theta) := \lceil \theta/\bar{k} \rceil,
  ~~~~~ ~~~~~ ~\:\:
  \ell(\theta) := k + (\theta \bmod \bar{k})
  ~\text{ for }~
  \theta \in \cC_1,
  \\
  &
  (- \bm x_{\ell(\theta)} - \hat{s}_{j(\theta)} \bm x_{j(\theta)} )^\top \bm y \le 0
  ~\text{ with }~
  j(\theta) := \lceil (\theta - k\bar{k})/\bar{k} \rceil,
  \ell(\theta) := k + (\theta \bmod \bar{k})
  ~\text{ for }~
  \theta \in \cC_2.
 \end{align*}
 Then, 
 the lower and the upper truncation points
 in Theorem \ref{theo:LeeEtAl} are written as
 \begin{subequations}
 \begin{align}
  \label{eq:vneg2}
  \!\!\!\!\!
  V^-(\bm A, \bm b)
  &=
  \max \left\{
  \max_{
  \substack{
  \theta \in [2k\bar{k}] \\
  \rho_{j(\theta)} + \bm x_{\ell(\theta)}^\top \bm \chi_\theta < 0
  }
  }
  \frac{\kappa_{j(\theta)} + \bm x_{\ell(\theta)}^\top \bm \xi_\theta}{\rho_{j(\theta)} + \bm x_{\ell(\theta)}^\top \bm \chi_\theta},
  ~
  \max_{
  \substack{
  j \in \cS\\
  \rho_j < 0  
  }
  }
  \frac{\kappa_j}{\rho_j}
  \right\}
  + \bm \eta^\top \bm y,
  \\
  \label{eq:vpos2}
  \!\!\!\!\!
  V^+(\bm A, \bm b)
  &=
  \min\left\{
  \min_{
  \substack{
  \theta \in [2k\bar{k}]\\
  \rho_{j(\theta)} + \bm x_{\ell(\theta)}^\top \bm \chi_\theta > 0
  }
  }
  \frac{\kappa_{j(\theta)} + \bm x_{\ell(\theta)}^\top \bm \xi_\theta}{\rho_{j(\theta)} + \bm x_{\ell(\theta)}^\top \bm \chi_\theta},
  ~
  \min_{
  \substack{
  j \in \cS\\
  \rho_j > 0
  }
  }
  \frac{\kappa_j}{\rho_j}
  \right\}
  + \bm \eta^\top \bm y,
 \end{align}
 \end{subequations}
 where,
 for
 $j \in [D]$
 and
 $\theta \in [2 k \bar{k}]$,
 \begin{align*}
  \kappa_j := \hat{s}_j \bm x_j^\top \bm y,
  ~
  \rho_j := - \hat{s}_j \bm x_j^\top \bm c,
  ~
  \bm \xi_\theta := \mycase{
  - \bm y & \text{if } \theta \in \cC_1, \\
  \phantom{-} \bm y & \text{if } \theta \in \cC_2,
  }
  ~
  \bm \chi_\theta := \mycase{
  \phantom{-} \bm c & \text{if } \theta \in \cC_1, \\
  - \bm c & \text{if } \theta \in \cC_2,
  }
 \end{align*}
with
$\bm c = \frac{\Sigma \bm \eta}{\bm \eta^\top \Sigma \bm \eta}$
as defined before. 
 Furthermore,
 let
\begin{align*}
 a_{j(\theta)}^+ \!:=\! \! \sum_{i | \xi_\theta^i > 0} \! x_{j(\theta)}^i \xi_\theta^i,
 ~
 a_{j(\theta)}^- \!:=\! - \! \sum_{i | \xi_\theta^i < 0} \! x_{j(\theta)}^i \xi_\theta^i,
 ~
 b_{j(\theta)}^+ \!:=\! \! \sum_{i | \chi_\theta^i > 0} \! x_{j(\theta)}^i \chi_\theta^i,
 ~
 b_{j(\theta)}^- \!:=\! - \! \sum_{i | \chi_\theta^i < 0} \! x_{j(\theta)}^i \chi_\theta^i,
\end{align*} 
where
 $\xi_\theta^i$
 and 
 $\chi_\theta^i$
 is the $i^{\rm th}$ element of
 $\bm \xi_\theta$
 and 
 $\bm \chi_\theta$,
 respectively.
 
 For each of the selected feature 
 $j \in \cS$,
 consider a tree structure as depicted in \figurename~\ref{fig:tree} 
 which only has a set of nodes 
 corresponding to each of the unselected features $\ell \in \bar{\cS}$.
 Considering a tree for a selected feature $j \in \cS$, 
 if a node corresponding to 
 $\ell(\theta)$, 
 $\theta \in \{\theta | j(\theta) = j\}$ 
 satisfies 
 \begin{align}
  \label{eq:v-neg-main}
  V^{-}_{\rm best}
  \ge
  \mycase{
  -
  \frac{
  \kappa_{j(\theta)} - a_{\ell(\theta)}^-
  }{
  |\rho_{j(\theta)} - b_{\ell(\theta)}^-|
  }  + \bm \eta^\top \bm y
  &
  \text{if }
  \rho_{j(\theta)} + b_{\ell(\theta)}^+ < 0, 
  \\
  -
  \frac{
  \kappa_{j(\theta)} - a_{\ell(\theta)}^-
  }{
  \max
  \{
  |\rho_{j(\theta)} - b_{\ell(\theta)}^-|, |\rho_{j(\theta)} + b_{\ell(\theta)}^+|
  \}
  } + \bm \eta^\top \bm y
  &
  \text{otherwise}.
  }
 \end{align}
 then 
 all the constraints indexed by
 $\theta^\prime$
 such that
 $\ell(\theta^\prime)$
 is a descendant of
 $\ell(\theta)$ 
 in the tree
 are guaranteed to have no influence on the lower truncation point
 $V^-(\bm A, \bm b)$,
 where
 $V^-_{\rm best}$
 is the current maximum of
 $V^-(\bm A, \bm b)$
 in \eq{eq:vneg2}. 
 Similarly,
 if 
 \begin{align}
  \label{eq:v-pos-main}
  V^{+}_{\rm best}
  \le
  \mycase{
  \frac{
  \kappa_{j(\theta)} - a_{\ell(\theta)}^-
  }{
  |\rho_{j(\theta)} + b_{\ell(\theta)}^+|
  } + \bm \eta^\top \bm y
  &
  \text{if }
  \rho_{j(\theta)} - b_{\ell(\theta)}^- > 0, 
  \\
  \frac{
  \kappa_{j(\theta)} - a_{\ell(\theta)}^-
  }{
  \max
  \{
  |\rho_{j(\theta)} - b_{\ell(\theta)}^-|, |\rho_{j(\theta)} + b_{\ell(\theta)}^+|
  \}
  } + \bm \eta^\top \bm y
  &
  \text{otherwise}.
  }
 \end{align}
 then 
 all the constraints indexed by
 $\theta^\prime$
 such that
 $\ell(\theta^\prime)$
 is a descendant of
 $\ell(\theta)$ 
 in the tree
 are guaranteed to have no influence on the upper truncation point
 $V^+(\bm A, \bm b)$,
 where
 $V^+_{\rm best}$
 is the current minimum of
 $V^+(\bm A, \bm b)$
 in \eq{eq:vpos2}. 
\end{theo}

The proof is presented in Supplementary Appendix \ref{app:proofs}.
Note that 
\eq{eq:v-neg-main} 
and
\eq{eq:v-pos-main}
can be evaluated by using information available at the node $\ell(\theta)$. 
If the conditions in
\eq{eq:v-neg-main}
or 
\eq{eq:v-pos-main}
are satisfied,
we can stop searching the tree
because it is guaranteed that any constraints
indexed by
$\theta^\prime$
such that
$\ell(\theta^\prime) \in De(\ell(\theta))$
do not have any influences on the truncation point,
and hence does not affect the sampling distribution for PSI.
%

%% file: sec4.tex
\section{Experiments}
\label{sec:exp}
%
\subsection{Experiments on synthetic data}
First, 
we checked the validity of our post-selection inference algorithm
for high-order interaction models
by using synthetic data. 
In the synthetic data experiments,
we compared our approach
(denoted as {\tt PSI}: Post-Selection Inference)
with
ordinary least-squares method ({\tt OLS})
and
data-splitting method ({\tt Split}).
In data splitting method,
the data set was randomly divided into two equal-sized subsets,
and one of them was used for feature selection,
while the other was used for statistical inference on post-regression model. 

The synthetic data was generated from
$
\bm y = \bm X \bm \beta + \bm \veps,
~
\bm \veps \sim N(\bm{0}, \sigma^2 \bm I),
$
where
$\bm y \in \RR^n$
is the response vector, 
$\bm X \in \{0, 1\}^{n \times D}$
is
the design matrix,
and
$\bm \veps \in \RR^n$
is the Gaussian noise vector.
Here,
we did not actually compute the extremely wide design matrix $\bm X$
because it has exponentially large number of columns. 
Instead, 
we generated a
random binary matrix 
$\bm Z \in \{0, 1\}^{n \times d}$
and each expanded high-order interaction feature 
$x^i_j$
was generated
from the $i^{\rm th}$ row of $\bm Z$
only when it was needed. 
For simplicity and computational efficiency,
we assumed that the covariates (hence interaction features as well) are binary,
and the sparsity rate $\eta \in [0, 1]$
(the rate of zeros in the entries of $\bm Z$)
was changed to see how sparsity is useful for efficient computation. 
As the baseline,
the rest of the parameters were set as
$n=100$,
$d=100$,
$\eta = 0.5$,
$\sigma = 0.1$,
$r=3$,
$k=10$,
and significance level
$\alpha = 0.05$. 
%
%

\begin{figure}[!h]
  \centering
  \begin{tabular}{ccc}
   \includegraphics[width=0.33\textwidth]{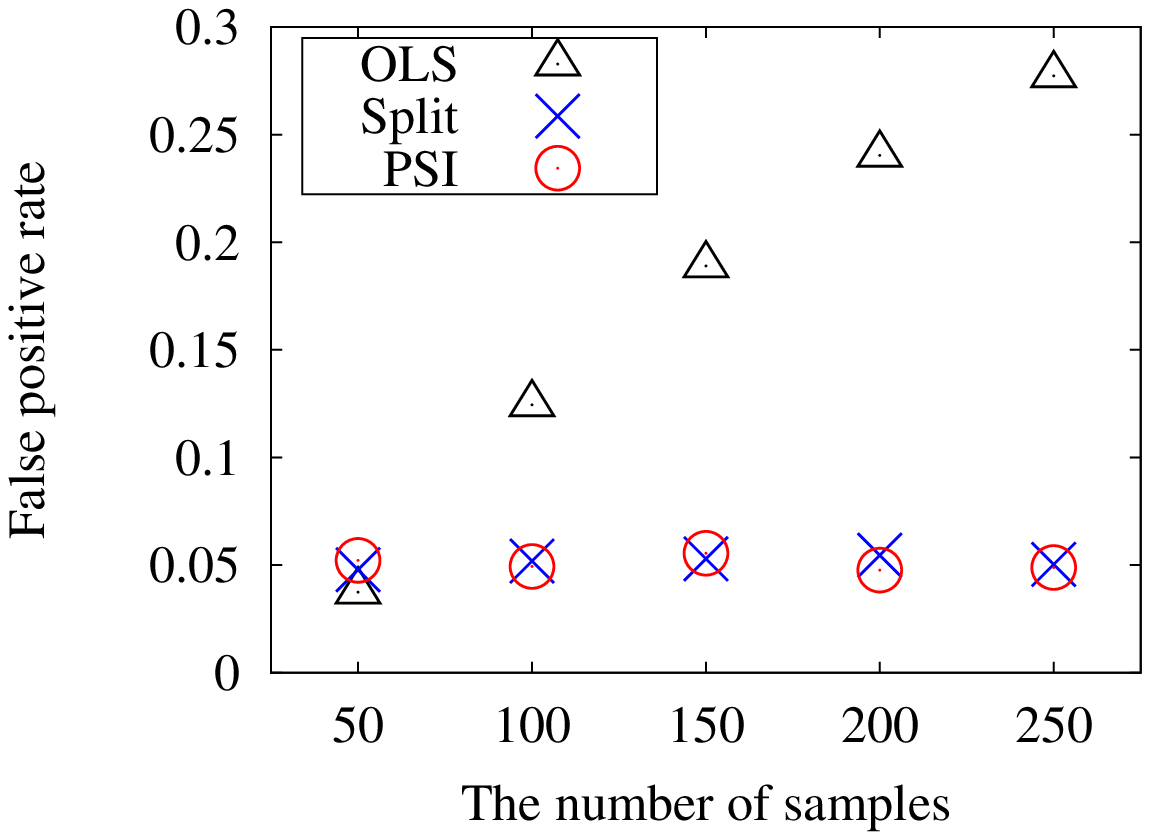} &
   \includegraphics[width=0.33\textwidth]{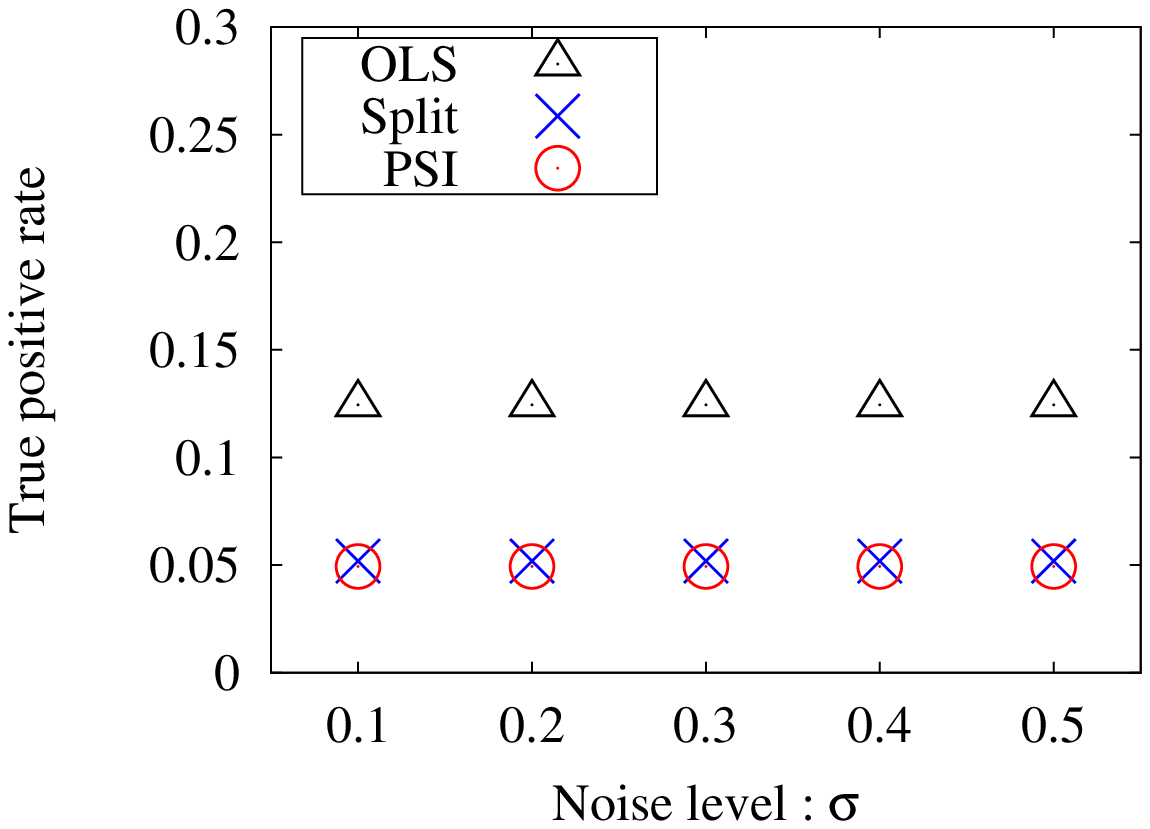} &
   \includegraphics[width=0.33\textwidth]{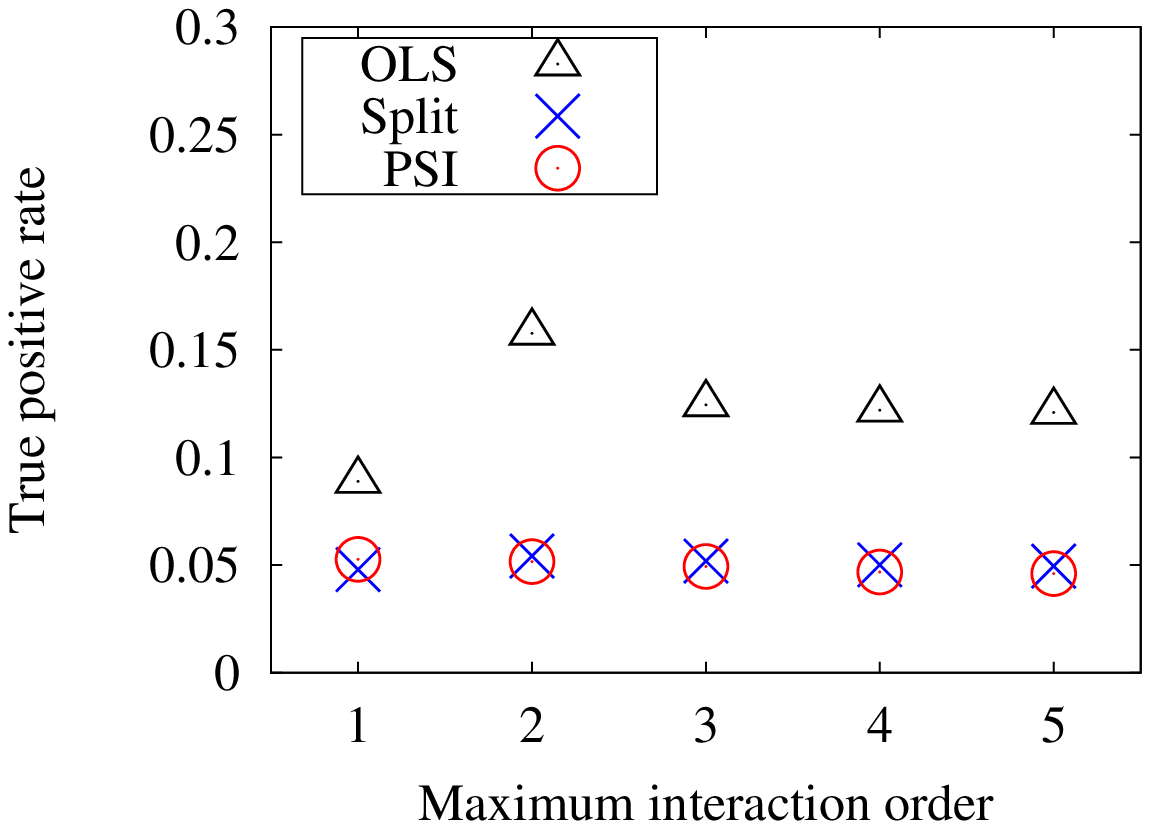} \\
   (a) $n \in \{50,\cdots,250\}$ &
   (b) $\sigma \in \{0.1,\cdots,0.5\}$ &
   (c) $r \in \{1,\cdots,5\}$
  \end{tabular}
  \caption{False positive rates of three methods {\tt PSI}, {\tt OLS} and {\tt Split}.}
  \label{fig:false_positive_rate}
  \begin{tabular}{ccc}
   \includegraphics[width=0.33\textwidth]{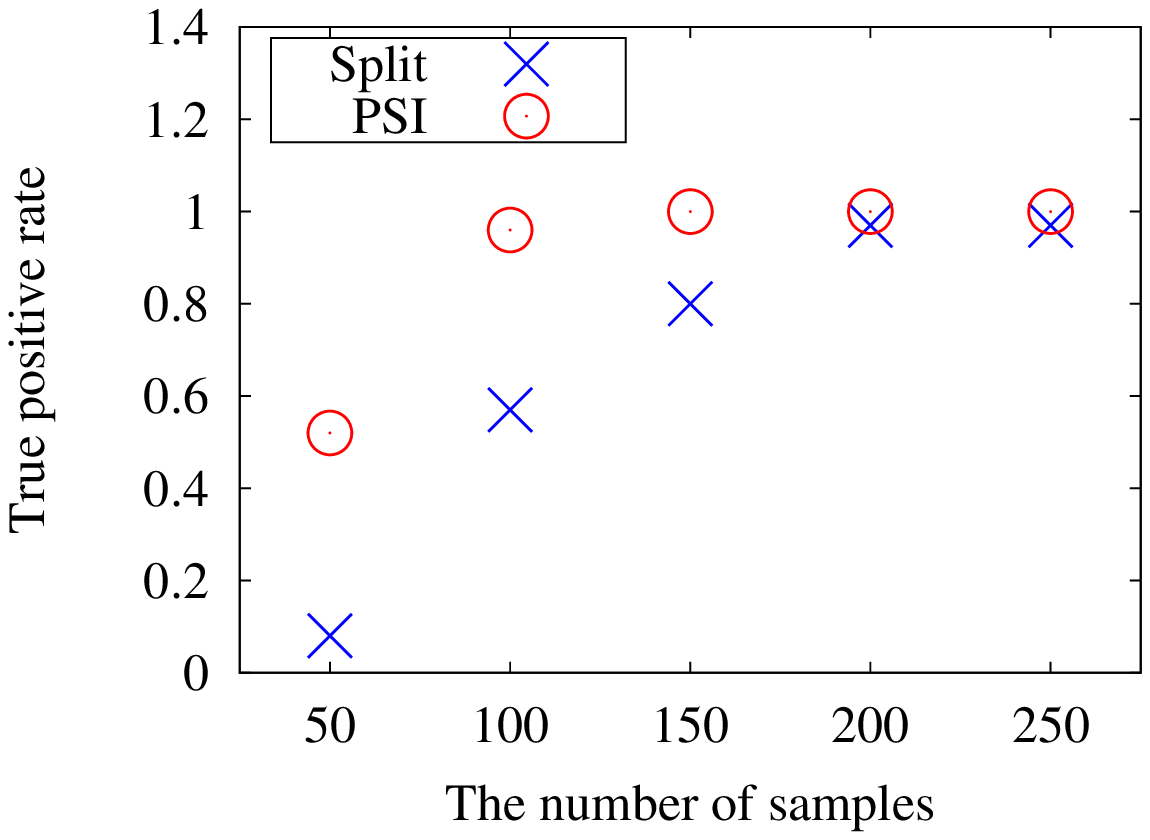} &
   \includegraphics[width=0.33\textwidth]{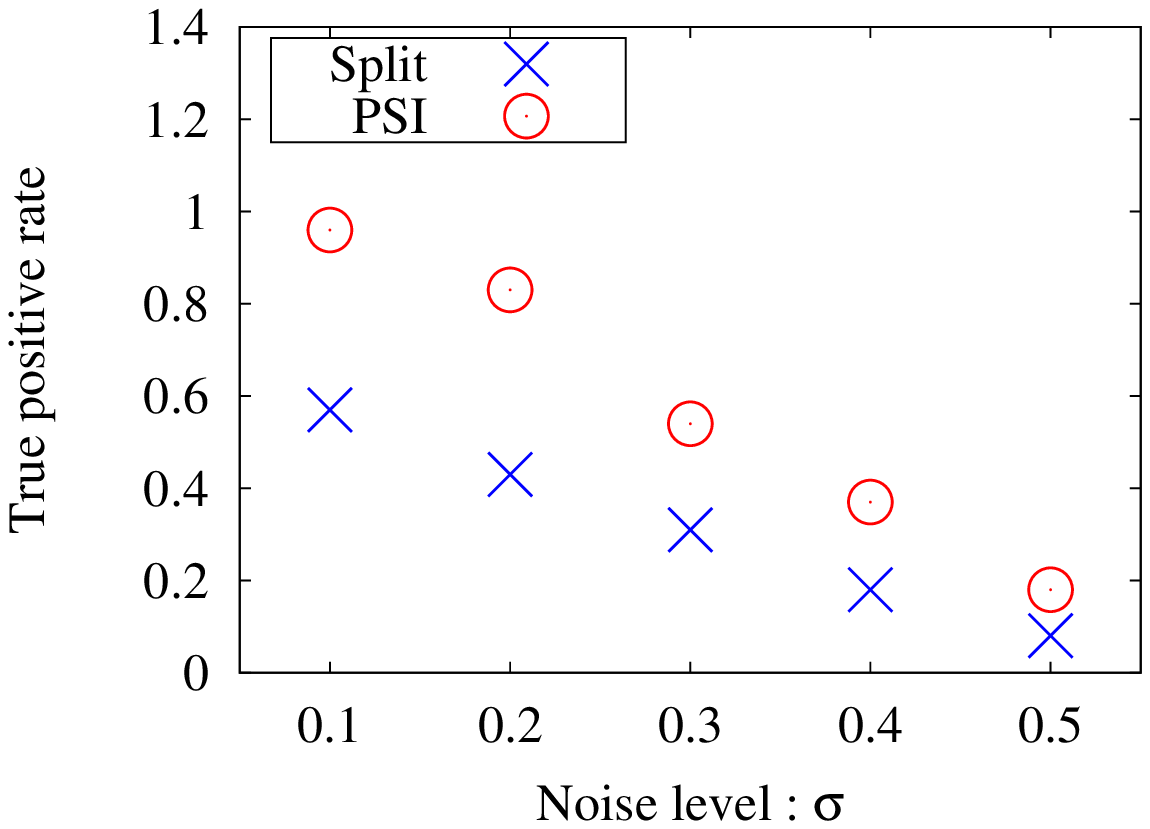} &
   \includegraphics[width=0.33\textwidth]{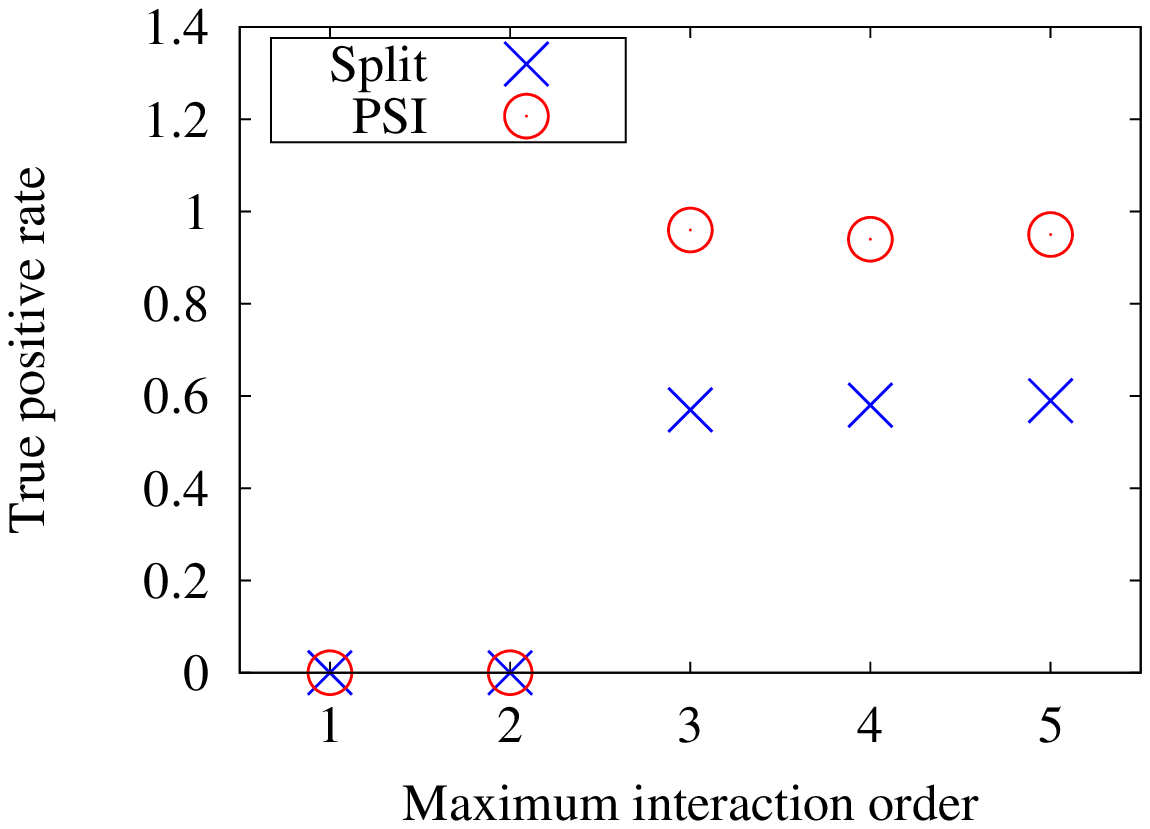} \\
   (a) $n \in \{50,\cdots,250\}$ &
   (b) $\sigma \in \{0.1,\cdots,0.5\}$ &
   (c) $r \in \{1,\cdots,5\}$
  \end{tabular}
  \caption{True positive rates of {\tt PSI} and {\tt Split}.}
  \label{fig:true_positive_rate}
\end{figure}

\paragraph{False positive rate control}
First,
to check whether the methods can properly control the desired false positive rate,
we generated data sets
with
$\bm \beta = \zero$
and see how many false positives would be reported by each of the three methods.
\figurename~\ref{fig:false_positive_rate}
shows the false positive rate
defined as
$k^\prime/k$
where
$k^\prime$
is the number of features reported as positives in post-regression models.
The plots in 
\figurename~\ref{fig:false_positive_rate}
are the averages over 1000 different trials
for various sample size $n \in \{50,\cdots,250\}$, noise level $\sigma \in \{0.1,\cdots,0.5\}$, and maximum interaction order $r \in \{1,\cdots,5\}$. 
As expected,
{\tt OLS}
could not properly control the false positive rates
because statistical inferences on the post-regression models would be positively biased 
when the features were selected by using the same data set.
On the other hand,
{\tt PSI} and {\tt Split}
could keep the false positive rates as desired $\alpha = 0.05$ level. 

\paragraph{True positive rate comparison}
Next,
we compared true positive rates of 
{\tt PSI} and {\tt Split}.
%
We set all the coefficient as 0
except 
$\beta_j = 1$
for
$j^{\rm th}$
feature 
corresponding to the $3^{\rm rd}$ order interaction term 
$z_1 z_2 z_3$. 
In 
\figurename~\ref{fig:true_positive_rate}, 
we report the true positive rates for various parameters,
which are defined as the number of times when the feature $z_1 z_2 z_3$ was detected as positive 
over the number of entire trials.
In almost all cases,
we see that 
{\tt PSI}
has larger true positive rates 
than 
{\tt Split}.
This is reasonable because 
the sample size used in the post-selection inference in the latter was half of the former. 

\paragraph{Computational efficiency}
Finally,
we demonstrate the computational efficiency of the proposed method.
In Tables~\ref{tb:exp_cond1} and \ref{tb:exp_cond2},
we show the average computation times on 10 trials in seconds 
and $1-$pruning rates for tree traversing with various those parameters.
Although the computation time increases
with
$d, ~r$ and $1 - \eta$,
the computation times were still in acceptable range. 

\begin{table}[!h]
 \vspace*{-5mm}
 \centering
 \caption{Computation times [sec] and $1-$pruning rates ($r = 3$).
}
  \label{tb:exp_cond1}
 \begin{scriptsize}
 \begin{tabular}{r|c|c|c|c} \hline
$d$	&	Time ($\eta = 0.9$)	&	Time ($\eta = 0.7$)	&	$1-$ pruning rate ($\eta = 0.9$)	&	$1-$ pruning rate ($\eta = 0.7$)	\\ \hline
100	&	~~0.01(0.002)	&	~~0.01(0.007)	&	$1.29\times10^{-2}(3.42\times10^{-3})$	&	$1.76\times10^{-2}(1.08\times10^{-2})$	\\
500	&	0.08(0.02)	&	0.16(0.11)	&	$9.71\times10^{-4}(2.63\times10^{-4})$	&	$1.79\times10^{-3}(1.30\times10^{-3})$	\\
1000	&	0.19(0.05)	&	0.52(0.37)	&	$2.91\times10^{-4}(9.77\times10^{-5})$	&	$7.11\times10^{-4}(5.49\times10^{-4})$	\\
5000	&	1.03(0.33)	&	3.36(2.92)	&	$1.18\times10^{-5}(3.73\times10^{-6})$	&	$3.67\times10^{-5}(3.28\times10^{-5})$	\\
10000	&	2.70(1.09)	&	9.50(9.20)	&	$4.12\times10^{-6}(1.45\times10^{-6})$	&	$1.39\times10^{-5}(1.32\times10^{-5})$	\\ \hline
 \end{tabular}
 \end{scriptsize}
  \caption{Computation times and $1-$pruning rates ($d = 5000$).}
  \label{tb:exp_cond2}
 \begin{scriptsize}
 \begin{tabular}{r|c|c|l|l} \hline
$r$	&	Time ($\eta = 0.9$)	&	Time ($\eta = 0.7$)	&	$1-$ pruning rate ($\eta = 0.9$)	&	$1-$ pruning rate ($\eta = 0.7$)	\\ \hline
1	&	~~0.03(0.001)	&	~~0.03(0.001)	&	$1(0)$	&	$1(0)$	\\
2	&	1.01(0.32)	&	2.59(2.23)	&	$1.96\times10^{-2}(6.23\times10^{-3})$	&	$5.20\times10^{-2}(4.54\times10^{-2})$	\\
3	&	1.05(0.33)	&	3.24(2.78)	&	$1.18\times10^{-5}(3.73\times10^{-6})$	&	$3.67\times10^{-5}(3.28\times10^{-5})$	\\
4	&	1.04(0.33)	&	3.98(3.00)	&	$9.47\times10^{-9}(2.99\times10^{-9})$	&	$4.03\times10^{-8}(3.39\times10^{-8})$	\\
5	&	1.03(0.32)	&	4.17(3.35)	&	$9.48\times10^{-12}(2.99\times10^{-12})$	&	$4.04\times10^{-11}(3.39\times10^{-11})$	\\ \hline
 \end{tabular}
 \end{scriptsize}
  \caption{The number of features reported as positives and computation times [sec].}
 \label{tb:result_real_data}
 \begin{scriptsize}
 \begin{tabular}{c|c|c|c||c|c|c|D{.}{.}{-1}} \hline
	\multirow{2}{*}{Dataset} & \multicolumn{3}{c||}{\tt Split} & \multicolumn{3}{c|}{\tt PSI} &	\multicolumn{1}{c}{time of} \\ \cline{2-7}
	 & 1st	&	2nd & 3rd & 1st & 2nd & 3rd & \multicolumn{1}{c}{\tt PSI} \\ \hline
	Communities\&Crime09 ($d=253$) & 1 &  &  &   2 &  &  & 1.48 \\
	Communities\&Crime11 ($d=289$) &  & 2 &  &   3 & 2 & 1 & 1.46 \\
	BlogFeedback ($d=339$) &  & 2 & 1 &   4 & 12 & 13 & 2.17 \\
	SliceLocalization ($d=769$) &  & 2 &   28 &    & 1 & 29 & 26.33 \\
	UJIIndoorLoc ($d=1053$) & 3 & 4 &  &   5 & 4 & 1 & 4.29 \\ \hline
 \end{tabular}
 \end{scriptsize}
\end{table}

  \subsection{Experiments on real data}
Here we show the statistical power of {\tt PSI} and {\tt Split} in real data.
Since the true positive features in real data are unknown,
we show the number of features reported as positives in post-regression models
assuming that these two methods can properly control false positive rates
as we confirmed in synthetic data experiments. 
We obtained datasets from UCI data repository, which listed in the first column of Table~\ref{tb:result_real_data}.
Continuous covariates in the original datasets were first standardized to have the mean zero and the variance one,
and then represented the covariate by two binary variables,
each of which indicates whether the value is greater than 1 or the value is smaller than -1. 
We estimated the $\sigma$ in the same way as \cite{Lee14a}.
We set the maximum interaction order as $r=3$ and the number selected features by marginal screening as $k = 30$.
For simplicity and computational efficiency, we randomly sampled 1000 instances from each dataset.
Table~\ref{tb:result_real_data} shows the number of features reported as positives on {\tt PSI} and {\tt Split}.
In almost all cases,
{\tt PSI}
found more positive features than {\tt Split},
while the computational cost of
{\tt PSI}
were still in acceptable range. 

%% file: sec5.tex
\section{Conclusions}
In this paper we proposed an efficient PSI on high-order interaction models with
marginal screening-based pre-feature selection.
Our key idea is 
to derive a pruning condition
of the tree 
that quickly identifies a set of unrelated features
with PSI. 
The experimental results indicated that
the proposed method allows us to reliably identify statistically
significant high-order interaction features
with reasonable computational cost.

%% file: appA.tex
\section{Proof of Theorem~\ref{theo:main}}
\label{app:proofs}
In this appendix we prove Theorem~\ref{theo:main}. 

\begin{proof}[Proof of Theorem~\ref{theo:main}]
 We only show the lower truncation point part of the Theorem.
 Consider an arbitrary pair 
 $(\theta, \theta^\prime) \in [2 k \bar{k}]^2$
 such that 
 $j(\theta) = j(\theta^\prime)$
 and
 $\ell(\theta^\prime) \in De(\ell(\theta))$. 
 We first note that
 the fact that
 $x^i_{\ell(\theta^\prime)} \le x^i_{\ell(\theta)}$
 for all $i \in [n]$
 indicates that 
 \begin{align}
  \label{eq:prf-1}
  0 \le a^+_{\ell(\theta^\prime)} \le a^+_{\ell(\theta)},
  ~
  0 \le a^-_{\ell(\theta^\prime)} \le a^-_{\ell(\theta)},  
  ~
  0 \le b^+_{\ell(\theta^\prime)} \le b^+_{\ell(\theta)},
  ~
  0 \le b^-_{\ell(\theta^\prime)} \le b^-_{\ell(\theta)}
 \end{align}

 (i) First 
 we prove 
 the first case in
 \eq{eq:v-neg-main}.
 Using the relations in
 \ref{eq:prf-1},
 we have
 \begin{align}
  \label{eq:prf-2}
  \rho_{j(\theta^\prime)} + \bm x_{\ell(\theta^\prime)}^\top \bm \chi_{\theta^\prime}
  =
  \rho_{j(\theta^\prime)} + b^+_{j(\theta^\prime)} - b^-_{j(\theta^\prime)}
  \le
  \rho_{j(\theta)} + b^+_{j(\theta)},
 \end{align}
 where we used
 $\rho_{j(\theta^\prime)} = \rho_{j(\theta)}$.
 Next, when
 $\rho_{j(\theta^\prime)} + \bm x_{\ell(\theta^\prime)}^\top \bm \chi_{\theta^\prime} < 0$,
 \begin{align}
  \label{eq:prf-3}
  \frac{
  \kappa_{j(\theta^\prime)} + \bm x_{\ell(\theta^\prime)}^\top \bm \xi_{\theta^\prime}
  }{
  \rho_{j(\theta^\prime)} + \bm x_{\ell(\theta^\prime)}^\top \bm \chi_{\theta^\prime}
  }
  = 
  \frac{
  \kappa_{j(\theta^\prime)} + a^+_{\ell(\theta^\prime)} - a^-_{\ell(\theta^\prime)}
  }{
  \rho_{j(\theta^\prime)} + b^+_{\ell(\theta^\prime)} - b^-_{\ell(\theta^\prime)}
  }
  \le 
  -
  \frac{
  \kappa_{j(\theta^\prime)} - a^-_{\ell(\theta^\prime)}
  }{
  |\rho_{j(\theta^\prime)} - b^-_{\ell(\theta^\prime)}|
  }
  \le 
  -
  \frac{
  \kappa_{j(\theta)} - a^-_{\ell(\theta)}
  }{
  |\rho_{j(\theta)} - b^-_{\ell(\theta)}|
  }, 
 \end{align}
 where
 we used the fact that
 the numerator is non-negative,
 and
 the denominator is non-positive
 in the left-most fraction. 
 From
 \eq{eq:prf-2}
 and 
 \eq{eq:prf-3},
 we have
 \begin{align*}
  \rho_{j(\theta)} + b^+_{j(\theta)} < 0
  \text{ and }
  -
  \frac{
  \kappa_{j(\theta)} - a^-_{\ell(\theta)}
  }{
  |\rho_{j(\theta)} - b^-_{\ell(\theta)}|
  }
  + \bm \eta^\top \bm y
  \le V^-_{\rm best}
  ~
  \Rightarrow
  ~
  \frac{
  \kappa_{j(\theta^\prime)} + \bm x_{\ell(\theta^\prime)}^\top \bm \xi_{\theta^\prime}
  }{
  \rho_{j(\theta^\prime)} + \bm x_{\ell(\theta^\prime)}^\top \bm \chi_{\theta^\prime}
  }
  + \bm \eta^\top \bm y
  \le V^-_{\rm best},
 \end{align*}
 which proves the first case in
 \eq{eq:v-neg-main}.
 (ii) Next,
 we prove the second case of 
 \eq{eq:v-neg-main}.
 When we do not know the sign of the denominator
 $\rho_{j(\theta^\prime)} + \bm x_{\ell(\theta^\prime)}^\top \bm \chi_{\theta^\prime}$,
 we can obtain a slightly loose bound in the following form
 \begin{align}
  \label{eq:prf-4}
  \begin{split}
  \frac{
  \kappa_{j(\theta^\prime)} + \bm x_{\ell(\theta^\prime)}^\top \bm \xi_{\theta^\prime}
  }{
  \rho_{j(\theta^\prime)} + \bm x_{\ell(\theta^\prime)}^\top \bm \chi_{\theta^\prime}
  }
  = 
  \frac{
  \kappa_{j(\theta^\prime)} + a^+_{\ell(\theta^\prime)} - a^-_{\ell(\theta^\prime)}
  }{
  \rho_{j(\theta^\prime)} + b^+_{\ell(\theta^\prime)} - b^-_{\ell(\theta^\prime)}
  }
  &
  \le 
  -
  \frac{
  \kappa_{j(\theta^\prime)} - a^-_{\ell(\theta^\prime)}
  }{
  \max\{
  |\rho_{j(\theta^\prime)} - b^-_{\ell(\theta^\prime)}|, 
  |\rho_{j(\theta^\prime)} + b^+_{\ell(\theta^\prime)}|
  \}
  }
  \\
  &
  \le 
  -
  \frac{
  \kappa_{j(\theta)} - a^-_{\ell(\theta)}
  }{
  \max\{
  |\rho_{j(\theta)} - b^-_{\ell(\theta)}|, 
  |\rho_{j(\theta)} + b^+_{\ell(\theta)}|
  \}
  }.
  \end{split}
 \end{align}
 From \eq{eq:prf-4},
 \begin{align*}
  -
  \frac{
  \kappa_{j(\theta)} - a^-_{\ell(\theta)}
  }{
  \max\{
  |\rho_{j(\theta)} - b^-_{\ell(\theta)}|, 
  |\rho_{j(\theta)} + b^+_{\ell(\theta)}|
  \}
  }
  + \bm \eta^\top \bm y
  \le
  V^-_{\rm best}
  ~\Rightarrow~
  \frac{
  \kappa_{j(\theta^\prime)} + \bm x_{\ell(\theta^\prime)}^\top \bm \xi_{\theta^\prime}
  }{
  \rho_{j(\theta^\prime)} + \bm x_{\ell(\theta^\prime)}^\top \bm \chi_{\theta^\prime}
  }
  + \bm \eta^\top \bm y
  \le
  V^-_{\rm best},
 \end{align*}
 which proves the second case of 
 \eq{eq:v-neg-main}. 
 Combining
 (i) and (ii),
 we showed that,
 if \eq{eq:v-neg-main} is satisfied for a certain $\theta$, 
 then any constraints indexed by
 $\theta^\prime$
 such that
 $j(\theta^\prime) = j(\theta)$
 and
 $\ell(\theta^\prime) \in De(\ell(\theta))$
 are guaranteed to have no influence on the lower truncation point $V^-(\bm A, \bm b)$.
 The upper truncation point part of the Theorem can be shown similarly. 
\end{proof}

%% file: paper.bbl
\begin{thebibliography}{10}
\bibitem{Lee14a}
J.~D. Lee and J.~E. Taylor.
\newblock Exact post model selection inference for marginal screening.
\newblock In {\em Advances in Neural Information Processing Systems}, 2014.

\bibitem{Lee15a}
J.~D. Lee, D.~L. Sun, Y.~Sun, and J.~E. Taylor.
\newblock Exact post-selection inference with applications to the {LASSO}.
\newblock {\em arXiv:1311.6238v5}, 2015.

\bibitem{manolio2006genes}
Teri~A Manolio and Francis~S Collins.
\newblock Genes, environment, health, and disease: facing up to complexity.
\newblock {\em Human heredity}, 63(2):63--66, 2006.

\bibitem{cordell2009detecting}
Heather~J Cordell.
\newblock Detecting gene--gene interactions that underlie human diseases.
\newblock {\em Nature Reviews Genetics}, 10(6):392--404, 2009.

\bibitem{leeb2005model}
Hannes Leeb and Benedikt~M P{\"o}tscher.
\newblock Model selection and inference: Facts and fiction.
\newblock {\em Econometric Theory}, 21(01):21--59, 2005.

\bibitem{leeb2006can}
Hannes Leeb and Benedikt~M P{\"o}tscher.
\newblock Can one estimate the conditional distribution of post-model-selection
  estimators?
\newblock {\em The Annals of Statistics}, pages 2554--2591, 2006.

\bibitem{Fan08a}
J.~Fan and J.~Lv.
\newblock Sure independence screening for ultrahigh dimensional feature space.
\newblock {\em Journal of The Royal Statistical Society B}, 70:849--911, 2008.

\bibitem{Fan09a}
J.~Fan, R.~Samworth, and Y.~Wu.
\newblock Ultrahigh dimensional feature selection: beyond the linear model.
\newblock {\em The Journal of Machine Learning Research}, 10:2013--2038, 2009.

\bibitem{Fan10a}
J.~Fan and R.~Song.
\newblock Sure independence screening in generalized linear models with
  np-dimensionality.
\newblock {\em Annals of Statistics}, 38:3567--3604, 2010.

\bibitem{Genovese12a}
C.~R. Genovese, J.~Jin, L.~Wasserman, and Z.~Yao.
\newblock A comparison of the lasso and marginal regression.
\newblock {\em The Journal of Machine Learning Research}, 13:2107--2143, 2012.

\bibitem{Choi10a}
N.H. Choi, W.~Li, and J.~Zhu.
\newblock Variable selection with the strong heredity constraint and its oracle
  property.
\newblock {\em Journal of the American Statistical Association}, 105:354--364,
  2010.

\bibitem{hao2014interaction}
Ning Hao and Hao~Helen Zhang.
\newblock Interaction screening for ultrahigh-dimensional data.
\newblock {\em Journal of the American Statistical Association},
  109(507):1285--1301, 2014.

\bibitem{Bien13a}
J.~Bien, J.~E. Taylor, and R.~Tibshirani.
\newblock A {LASSO} for hierarchical interactions.
\newblock {\em Journal of The Royal Statistical Society B}, 41:1111--1141,
  2013.

\bibitem{hamalainen2014statistically}
Wilhelmiina H{\"a}m{\"a}l{\"a}inen and Geoff Webb.
\newblock Statistically sound pattern discovery.
\newblock In {\em Proceedings of the 20th ACM SIGKDD international conference
  on Knowledge discovery and data mining}, pages 1976--1976. ACM, 2014.

\bibitem{Saigo06a}
H.~Saigo, T.~Uno, and K.~Tsuda.
\newblock Mining complex genotypic features for predicting hiv-1 drug
  resistance.
\newblock {\em Bioinformatics}, 24:2455–--2462, 2006.

\bibitem{Kudo05a}
T.~Kudo, E.~Maeda, and Y.~Matsumoto.
\newblock An application of boosting to graph classification.
\newblock In {\em Advances in Neural Information Processing Systems}, 2005.

\bibitem{Morishita02a}
S.~Morishita.
\newblock Computing optimal hypotheses efficiently for boosting.
\newblock {\em Lecture Notes in Computer Science}, 2281:471--481, 2002.

\bibitem{tusher2001significance}
Virginia~Goss Tusher, Robert Tibshirani, and Gilbert Chu.
\newblock Significance analysis of microarrays applied to the ionizing
  radiation response.
\newblock {\em Proceedings of the National Academy of Sciences},
  98(9):5116--5121, 2001.

\bibitem{dudoit2003multiple}
Sandrine Dudoit, Juliet~Popper Shaffer, and Jennifer~C Boldrick.
\newblock Multiple hypothesis testing in microarray experiments.
\newblock {\em Statistical Science}, pages 71--103, 2003.

\bibitem{terada2013statistical}
Aika Terada, Mariko Okada-Hatakeyama, Koji Tsuda, and Jun Sese.
\newblock Statistical significance of combinatorial regulations.
\newblock {\em Proceedings of the National Academy of Sciences},
  110(32):12996--13001, 2013.

\bibitem{lopez2015fast}
Felipe~Llinares L{\'o}pez, Mahito Sugiyama, Laetitia Papaxanthos, and Karsten~M
  Borgwardt.
\newblock Fast and memory-efficient significant pattern mining via permutation
  testing.
\newblock {\em arXiv preprint arXiv:1502.04315}, 2015.

\bibitem{berk2013valid}
Richard Berk, Lawrence Brown, Andreas Buja, Kai Zhang, Linda Zhao, et~al.
\newblock Valid post-selection inference.
\newblock {\em The Annals of Statistics}, 41(2):802--837, 2013.

\bibitem{lockhart2014significance}
Richard Lockhart, Jonathan Taylor, Ryan~J Tibshirani, and Robert Tibshirani.
\newblock A significance test for the lasso.
\newblock {\em Annals of statistics}, 42(2):413, 2014.
\end{thebibliography}
